\documentclass{article}

\usepackage{jmlr2e}
\usepackage[utf8]{inputenc} 
\usepackage[T1]{fontenc}    
\usepackage{hyperref}       
\usepackage{url}            
\usepackage{booktabs}       
\usepackage{amsfonts}       
\usepackage{nicefrac}       
\usepackage{microtype}      
\usepackage{graphicx}
\usepackage{tabularx}
\usepackage{xcolor}
\usepackage{listings}

\newcommand{\jefftojacob}[1]{{\color{red} JeffToJacob: #1}}
\renewcommand{\jefftojacob}[1]{}

\definecolor{light-gray}{gray}{0.95}
\newcommand{\code}[1]{\colorbox{light-gray}{\texttt{#1}}}

\usepackage{lastpage}
\jmlrheading{}{}{}{}{}{}{Jacob Schreiber}

\ShortHeadings{apricot: Submodular selection for data summarization}{Schreiber}

\begin{document} 

\title{apricot: Submodular selection for data summarization in Python}

\author{\name Jacob Schreiber \email jmschr@cs.washington.edu \\
    \addr Paul G. Allen School of Computer Science and Engineering, University of Washington, Seattle, WA 98195-4322, USA \\
\AND
\name Jeffrey Bilmes \email bilmes@uw.edu \\
    \addr Department of Electrical \& Computer Engineering, University of Washington, Seattle, WA 98195-4322, USA \\
\AND
\name William Stafford Noble \email william-noble@uw.edu \\
    \addr Department of Genome Science, University of Washington, Seattle, WA 98195-4322, USA \\
}

\editor{}

\maketitle

\begin{abstract}%
We present apricot, an open source Python package for selecting representative subsets from large data sets using submodular optimization. The package implements an efficient greedy selection algorithm that offers strong theoretical guarantees on the quality of the selected set. Two submodular set functions are implemented in apricot: facility location, which is broadly applicable but requires memory quadratic in the number of examples in the data set, and a feature-based function that is less broadly applicable but can scale to millions of examples. Apricot is extremely efficient, using both algorithmic speedups such as the lazy greedy algorithm and code optimizers such as numba. We demonstrate the use of subset selection by training machine learning models to comparable accuracy using either the full data set or a representative subset thereof. This paper presents an explanation of submodular selection, an overview of the features in apricot, and an application to several data sets. The code and tutorial Jupyter notebooks are available at \url{https://github.com/jmschrei/apricot}
\end{abstract}


\section{Introduction}
Recent years have seen a surge in the number of massive publicly available data sets across a variety of fields. Relative to smaller data sets, larger data sets offer higher coverage of important modalities that exist within the data, as well as examples of rare events that are nonetheless important. However, as data sets become larger, they risk containing redundant data. Indeed, \cite{cifar-dup} found almost 800 nearly identical images in the popular CIFAR-10 \citep{cifar} data set of images.

Several existing Python packages focus on selecting subsets of features \citep{scikit-learn, msurf}; however, we are not aware of packages that identify subsets of examples from large data sets. Submodular selection has emerged as a potential solution to this problem \citep{lin2009-submod-active-seq}, by providing a framework for selecting examples that minimize redundancy with each other \citep{speechrecognition}. Specifically, submodular functions are those that, for any two sets $X,Y$ satisfying $X \subseteq Y$ and any example $v \notin Y$, have the ``diminishing returns'' property $f(X \cup \{ v \}) - f(X) \geq f(Y \cup \{v\}) - f(Y)$.  A function is also monotone if $f(X) \leq f(Y)$. In the setting of selecting examples from a large data set (called the ``ground set''), a monotone submodular function operates on a subset of the ground set and, due to the diminishing returns property, returns a value that is inversely related to the redundancy. Finding the subset of examples that maximizes this value, subject to a cardinality constraint, is NP-hard. Fortunately, a greedy algorithm can find a subset whose objective value is guaranteed to be within a constant factor $(1 - e^{-1})$ of the optimal subset \citep{seminal}. While other toolkits implement general purpose submodular optimization \citep{krause-2010}, they do not focus on this setting. For a more thorough introduction to submodularity, we recommend \cite{fujishige2005,krause2014submodular}.

In this work, we describe apricot, a Python package that implements submodular functions for the purpose of summarizing large data sets. The implementation of these functions is extremely efficient due to algorithmic tricks, such as the accelerated (or lazy) greedy algorithm \citep{lazygreedy}, and code optimizers, such as numba \citep{numba}. These functions are implemented using the API of scikit-learn transformers, allowing them to be easily dropped into existing machine learning pipelines. Lastly, because far more submodular functions exist than those currently included, apricot can be easily extended to user-defined submodular functions simply by defining what the gain would be of adding an example to the growing subset. Apricot can be easily installed using \verb|pip install apricot-select|.

\section{Facility location and feature-based functions}

\begin{figure*}
    \center
    \includegraphics[width=\textwidth]{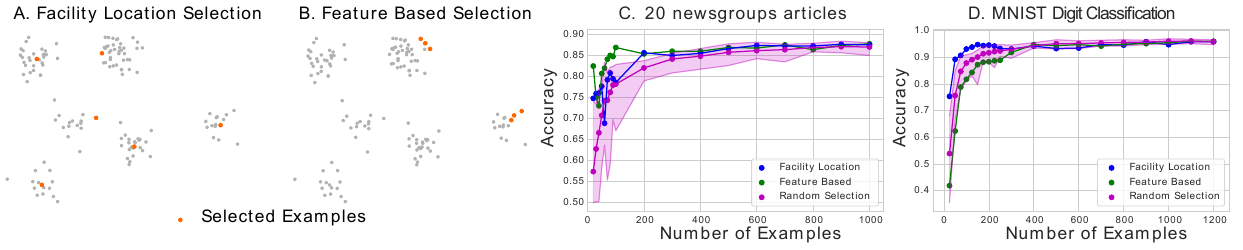}
    \caption{{\bf Example usage of submodular selection} (A-B) For the same data set, the examples that are selected using a (A) facility location function on correlations or a (B) feature-based function on the feature values directly are shown in orange, and the unselected examples are shown in gray. (C-D) Each dot represents the performance of a model trained on a subset of examples, selected either using a feature-based function (green), a facility location function (blue), or using random selection (magenta). The bounds show the minimum and maximum accuracy over 20 random selections. Models were evaluated on (C) classifying articles in the 20 newsgroups data set as being either related to space or medicine, and (D) classifying images of digits in the MNIST data set.}
    \label{fig:selection}
\end{figure*}

Facility location is a canonical submodular function parameterized by similarities between pairs of examples, such as correlations or cosine similarities. Specifically, facility location functions take the form
\begin{equation}
    f(X) = \sum\limits_{y \in Y} \max\limits_{x \in X} \phi(x, y)
\end{equation}
where $Y$ is a data set, $X$ is the selected subset where $X \subseteq Y$, $x$ and $y$ are examples in that data set, $\phi$ is the similarity function, and $\phi(x, y)$ is the similarity between the examples. When facility location functions are maximized, they tend to choose examples that represent the space of the data well. This can be visually seen on a toy data set of examples drawn from six distinct Gaussian distributions (Figure~\ref{fig:selection}A). Importantly, the similarity matrix must have non-negative entries, and larger values indicate greater similarity. Facility location functions can be called in apricot using the \code{FacilityLocationSelection} class.

However, a limitation of facility location functions is that, because they operate on a similarity matrix, they require memory that is quadratic with the number of examples, making them infeasible for large data sets. Although apricot allows one to pass in a sparse matrix, where the missing values are assumed to have similarities of zero, this requires knowing beforehand which entries in the matrix can be removed, and not all data sets have an underlying sparse similarity structure to them.

Feature-based selection (\citep{kirchhoff2014-submodmt} and \cite[Sec-3.1]{bilmes-dsf-arxiv-2017}) is an alternative approach that operates directly on feature values of examples. Maximizing a feature-based function involves selecting examples that differ in which feature values have high magnitudes. A feature-based function is thus a good choice when the features each represent some quality of the underlying data, where a higher value indicates an greater prevalence of that quality and a zero value indicates that the quality is entirely missing from the given example. An important restriction of feature-based selection approaches is that the feature values cannot be negative.
More formally, feature-based functions take the form
\begin{equation}
    f(X) = \sum\limits_{d=1}^{D} w_{d} \phi \left( \sum\limits_{x \in X} x_{d} \right),
\end{equation}
where $X \subseteq Y$, $Y$ is a data set, $d$ is a single feature in the data set, $D$ is the number of features in the data set, $w_d$ is the importance weight of feature $d$, $x$ is an example, $x_{d}$ is the value of feature $d$ in example $x$, and $\phi$ is a monotone concave function, such as \code{sqrt} or \code{log}.  The concave function provides the diminishing returns property. Feature-based functions can be called in apricot using the \code{FeatureBasedSelection} class.

Note that not all featurizations of data yield useful summaries when using a feature-based function. The assumption that each feature represents some quality of the data is reasonable for many data sets, such as when features correspond to word counts in a sentence or properties of a house, but not all data sets, such as pixels in an image.
For example, when applied to the same data set that the facility location function was, feature-based selection will choose examples with the highest feature values, which are not representative of the full data set (Figure~\ref{fig:selection}B). Fortunately, many transformations are available to construct features that are more amenable to feature-based selection, such as passing images through a neural network and performing selection on the internal representation \citep{lavania-autosum-sdm-2019}. Regardless, just as one should consider the similarity function to use for facility location functions, one should consider the feature representation used for feature-based functions.

A common application of submodular selection is to select a subset of data for faster training of machine learning models \citep{lin2009-submod-active-seq,kirchhoff2014-submodmt,wei2015-submodular-data-active}. The reasoning is that if the set reduces the redundancy in the data while preserving the diversity, then these examples will still be a good set to train on while yielding faster training. To illustrate this approach in apricot, we consider two examples: classifying articles as ``space'' vs ``medicine'' articles in the 20 newsgroups data set \citep{twentynewsgroups} and classifying digits in a subset of the MNIST data set \citep{mnist}, both using a tuned logistic regression model. For each example, we select subsets of increasing size using a feature-based function, a facility location function that uses squared correlations as similarities, and 20 iterations of random selection. The newsgroups features are TF-IDF transformed word counts, and the MNIST features are pixel saturation values. 

We observe that in the case of the 20 newsgroup data set the feature-based function appears to perform the best and achieves accuracy comparable to models trained on all 1000 training examples while using only 100 examples (Figure~\ref{fig:selection}C). However, consistent with our intuition that feature-based functions do not work well on all featurizations of data, we also observe that the corresponding feature-based functions do not perform as well on the MNIST example (Figure~\ref{fig:selection}D). This is because selecting examples with a diversity of pixels being saturated does not correspond to seeing relevant examples of different digits. In fact, high pixel values in rare pixels are likely to correspond to noise rather than meaningful signal. On the other hand, we observe that the facility location function applied to squared correlation coefficients performs well at this task, because examples of the same digit will overall display more similarly to each other than to other digits.

\section{The API}

Apricot implements submodular selection in the form of a scikit-learn transformer. Thus, parameters of the selection are passed into the constructor upon initialization, and the API consists of three functions: \code{fit}, which performs the selection and stores the identified ranking of examples, \code{transform}, which uses the stored ranking to select the subset of examples from the data set, and the \code{fit\_transform} function, which successively calls the \code{fit} and then the \code{transform} function on a data set. The parameters of the selection include the number of examples to select and the saturating function to use. An example of selecting a subset of size 100 from a large data set \verb|X| using a feature-based function with the \code{sqrt} saturating function would be
\begin{verbatim}
from apricot import FeatureBasedSelection
selection = FeatureBasedSelection(100, `sqrt', verbose=True)
X_subset = selection.fit_transform(X)
\end{verbatim}

We implement two optimization approaches. The na{\"i}ve greedy algorithm simply iterates through each example in the data set and calculates the gain in the objective function should it be added to the subset, and then selects the example with the largest gain. This process is repeated until the specified number of examples are chosen. However, discarding all knowledge of the ranking at each iteration is not optimal in terms of speed over the entire selection process. An extension of this selection procedure, called the accelerated greedy algorithm, can dramatically improve speed by ordering not-yet-chosen examples using a priority queue. Unfortunately, creating and maintaining the priority queue can be expensive and is not as easy to parallelize. While the lazy greedy algorithm generally leads to massive speed gains later on in the selection process relative to the na{\"i}ve greedy, the lazy greedy algorithm is not always faster than the parallelized na{\"i}ve algorithm early on. Thus, apricot offers a balance between the two where the initial rounds can be done in parallel using the na{\"i}ve greedy algorithm, and later rounds can employ the lazy greedy algorithm. The number of iterations to use before switching from the na{\"i}ve greedy algorithm to the lazy greedy algorithm is a parameter that is dataset specific. In practice, we have observed that between 50 and 500 generally works well for feature-based functions and between 1 and 50 generally works well for facility location functions.

The submodular selection process can be easily split into abstract parts. Accordingly, apricot offers a base selection class that maintains the priority queue, stores the sufficient statistics about the selected set for efficient computation of the gain, determines whether the na{\"i}ve greedy or the lazy greedy algorithm should be used for the next iteration of selection, implements the transform function, and handles user-defined initializations for the selection process. This base class is then inherited by the specific selection approaches, which implement code that calculates the gain of adding each example to the growing subset. This structure makes it simple for users to write their own submodular functions by inheriting from the base class and only having to write code that returns the gain of each example.

\section{Discussion}

A key challenge in any submodular selection is choosing the right input representation and the right submodular function. As we saw in the MNIST example, some functions work better than others on certain types of features. Therefore, careful consideration should be given to what function is used, and what features are used. In many cases, features that are not well suited for particular functions can be transformed using clever tricks. Alternatively, data can be clustered using mixture models and feature-based selection run on the posterior probabilities of each example, choosing examples that are close to each cluster center. While apricot does not directly implement these transformations, they would undoubtedly be valuable when used in conjunction with apricot.

\newpage
\acks{This work was supported by NSF IGERT award DGE-1258485 and by NIH award U01~HG009395.}

\bibliography{main}

\end{document}